\documentclass[conference]{IEEEtran}
\IEEEoverridecommandlockouts
\usepackage{cite}
\usepackage{amsmath,amssymb,amsfonts}
\usepackage{algorithmic}
\usepackage{graphicx}
\usepackage{textcomp}
\usepackage{xcolor}
\usepackage{float}
\usepackage{comment}
\def\BibTeX{{\rm B\kern-.05em{\sc i\kern-.025em b}\kern-.08em
    T\kern-.1667em\lower.7ex\hbox{E}\kern-.125emX}}
\begin{document}

\title{Exploring Narrative Clustering in Large Language Models: A Layerwise Analysis of BERT} 

\author{
\IEEEauthorblockN{Awritrojit Banerjee}
\IEEEauthorblockA{
\textit{CCN Group} \\
\textit{Pattern Recog. Lab.} \\
\textit{FAU Erlangen-Nürnberg}\\
Erlangen, Germany \\
awritrojit.banerjee@fau.de}
\and
\IEEEauthorblockN{Achim Schilling}
\IEEEauthorblockA{
\textit{Neuroscience Lab,} \\
\textit{University Hospital Erlangen,}\\ 
\textit{CCN Group} \\
\textit{Pattern Recog. Lab.} \\
\textit{FAU Erlangen-Nürnberg}\\
Erlangen, Germany \\
achim.schilling@fau.de\\
}
\and
\IEEEauthorblockN{Patrick Krauss}
\IEEEauthorblockA{
\textit{CCN Group} \\
\textit{Pattern Recog. Lab.} \\
\textit{FAU Erlangen-Nürnberg}\\
Erlangen, Germany \\
patrick.krauss@fau.de}
}


\maketitle

\begin{abstract}
This study investigates the internal mechanisms of BERT, a transformer-based large language model, with a focus on its ability to cluster narrative content and authorial style across its layers. Using a dataset of narratives developed via GPT-4, featuring diverse semantic content and stylistic variations, we analyze BERT's layerwise activations to uncover patterns of localized neural processing. Through dimensionality reduction techniques such as Principal Component Analysis (PCA) and Multidimensional Scaling (MDS), we reveal that BERT exhibits strong clustering based on narrative content in its later layers, with progressively compact and distinct clusters. While strong stylistic clustering might occur when narratives are rephrased into different text types (e.g., fables, sci-fi, kids’ stories), minimal clustering is observed for authorial style specific to individual writers. These findings highlight BERT's prioritization of semantic content over stylistic features, offering insights into its representational capabilities and processing hierarchy. This study contributes to understanding how transformer models like BERT encode linguistic information, paving the way for future interdisciplinary research in artificial intelligence and cognitive neuroscience.
\end{abstract}


\section{Introduction}

Artificial neural networks (ANNs), especially deep learning models, have achieved significant success in tasks involving sequential data, notably in natural language processing (NLP) and speech recognition \cite{lecun2015deep}. However, the extent to which these models replicate the localized processing observed in biological neural networks remains a subject of debate \cite{marcus2018deep}.

The transformer architecture, introduced by Vaswani \textit{et al.} \cite{vaswani2017attention}, has emerged as a promising model for bridging this gap. Transformers, particularly large language models (LLMs), are capable of performing complex cognitive tasks while offering the advantage of complete accessibility to their internal states and parameters, a feature unattainable in biological systems. This has led to their adoption as model systems in Cognitive Computational Neuroscience (CCN) \cite{Kriegeskorte2018}, a field dedicated to understanding cognitive processes by integrating artificial and biological neural systems \cite{Schilling2023, Stoewer2022, Stoewer2023a, Stoewer2023b, Surendra2023}. Through progressive refinements, these models can be made increasingly biologically plausible \cite{Pulvermuller2021, Gerum2020, Gerum2021, Gerum2023, Stoll2023}, providing insights into both the human brain and the principles underlying artificial intelligence \cite{Castelvecchi2016}.

Among transformers, the Bidirectional Encoder Representations from Transformers (BERT) \cite{devlin2019bert}, an encoder-only architecture, is particularly notable for its ability to process input sequences holistically. Its self-attention mechanism dynamically weights input tokens based on contextual importance, resembling the brain's ability to allocate resources to salient sensory inputs. This makes BERT a compelling framework for exploring the hypothesis that localized processing of sequential data can emerge in artificial neural networks.

Explainable AI (XAI) methods \cite{samek2017explainable} provide the tools to interpret and visualize the internal workings of such models. Techniques like Principal Component Analysis (PCA) and Multidimensional Scaling (MDS) enable dimensionality reduction, facilitating the identification of patterns and clusters in high-dimensional activation spaces. By applying these methods to BERT, we aim to investigate how narrative content and stylistic features manifest within its internal representations.

We constructed a dataset of texts featuring diverse authorial styles and narrative themes, using GPT-4 to ensure variability and consistency. By analyzing BERT's layerwise activations and projecting them into two-dimensional spaces using PCA and MDS, we reveal distinct clustering patterns based on narrative content, with progressively compact clusters in later layers. In contrast, clustering by authorial style is minimal, highlighting BERT's prioritization of semantic content over stylistic features.

These findings support the hypothesis that encoder-only transformers like BERT exhibit localized processing of sequential data, aligning with principles of hierarchical abstraction observed in the brain. The study underscores the potential of transformers as tools for cognitive modeling and highlights their implications for both neuroscience and artificial intelligence. In the following sections, we detail our methodology, present key results, and discuss the broader implications of our findings.
\\

\section{Methodology}

In this section, we detail the construction of our dataset, the application of a pre-trained language model to encode the data, and the techniques used to reduce dimensionality for visualization. Our objective is to generate a corpus that integrates both authorial style and narrative content, and then examine how these factors manifest within BERT’s internal activation patterns.

\subsection{Dataset Construction}

\subsubsection{Selection of Authors and Texts}
We began by selecting a set of authors and texts that collectively cover a wide spectrum of English literary styles and thematic domains. The chosen authors are, in order: William Shakespeare, Charles Dickens, Charlotte Brontë, Sir Arthur Conan Doyle, Edgar Allan Poe, George R. R. Martin, Hector Hugh Munro (Saki), William Sydney Porter (O. Henry), Dan Brown, and Jerome K. Jerome. Each author is represented by one text:
\begin{enumerate}
    \item \emph{Twelfth Night, or What You Will} (Shakespeare; a play that narrates a story)
    \item \emph{A Tale of Two Cities} (Dickens; realistic themes)
    \item \emph{Jane Eyre} (Brontë; passive and emotional)
    \item \emph{The Hound of the Baskervilles} (Doyle; thriller, realistic yet larger than life, analytical)
    \item \emph{The Tell-Tale Heart} (Poe; psychological, gritty)
    \item \emph{A Song of Ice and Fire: A Dance with Dragons} (Martin; fantasy with mature themes)
    \item \emph{The Wolves of Cernogratz} (Saki; seemingly straightforward yet enigmatic)
    \item \emph{The Last Leaf} (O. Henry; positively emotional)
    \item \emph{Angels and Demons} (Brown; analytical)
    \item \emph{Three Men in a Boat} (Jerome; humorous)
\end{enumerate}

This carefully curated selection ensures a balanced representation of major literary modes, encompassing diverse narrative styles, thematic complexities, emotional tones, and authorial voices.

\subsubsection{Neural Style Transfer}
From each of the 10 texts, we extracted the first few lines, yielding 10 base narratives. To introduce stylistic variability, we leveraged ChatGPT-4 to perform “neural style transfer” in the literary domain. Before initiating this process, ChatGPT-4 was instructed to act as an expert in Comparative Literature, specializing in pre-Victorian, Victorian, and modern English Literature, with an in-depth understanding of the selected authors.

For each of the 10 base narratives, we prompted ChatGPT-4 to rewrite it in the style of each of the other 9 authors. This procedure generated 10 differently stylized versions of each narrative (the original plus 9 new stylizations), resulting in 100 narratives in total. To ensure a suitable ratio of data points to embedding dimensions for subsequent analysis, we repeated this style-transfer step 10 times, ultimately producing a dataset of 1000 narratives. This expanded dataset is essential for enabling our chosen projection methods to operate on the column space (the activation space) rather than the row space.

\subsection{BERT: Representation Extraction}

\subsubsection{BERT as an Encoder-Only Transformer}
We employed the BERT-base-uncased model, a pre-trained encoder-only transformer architecture. BERT’s strength lies in its ability to capture contextual relationships between tokens in a bidirectional manner, without relying on recurrence or convolution. Its self-attention mechanism enables each token to attend directly to others, facilitating nuanced representations of semantic and syntactic features. By examining the internal activations of BERT, we aim to determine if and how it encodes the dual semantics of style and content present in our dataset.

\subsubsection{Tokenization and Embedding Retrieval}
Each of the 1000 narratives was tokenized using the BERT-base-uncased tokenizer, converting the text into a sequence of WordPiece tokens. We then passed each tokenized narrative through BERT, extracting the [CLS] token embedding from every layer. As BERT-base-uncased produces a 768-dimensional embedding per layer, and we consider all 13 layers (including the embedding layer), we obtained a data structure of shape $(13,\, 1000,\, 768)$. This encapsulates the layered representation of each narrative’s [CLS] token, capturing a progression of representations that BERT forms as information flows through its layers.

Our dataset now encodes two distinct kinds of semantics: narrative content (inherited from the original excerpts) and authorial style (imposed by the neural style transfer). As per our central hypothesis, we are expecting to see some localization of neural activations in  BERT’s internal representation space, but which of the semantics will influence the localization will be interesting to examine.

\subsection{Dimensionality Reduction and Projection}

To visualize and analyze the resulting representations, we employed two complementary dimensionality reduction methods: Principal Component Analysis (PCA) and Multidimensional Scaling (MDS).

\subsubsection{Principal Component Analysis (PCA)}
PCA identifies orthogonal directions in the data that capture maximal variance, projecting high-dimensional points into a lower-dimensional space. This approach reveals dominant axes of variation that may correspond to salient semantic distinctions. By ensuring that the number of narratives (1000) exceeds the embedding dimension (768), PCA can operate from the column space—our desired “activation space”—rather than being constrained by row-wise limitations.

\subsubsection{Multidimensional Scaling (MDS)}
This technique was used to reduce the dimensionality of the hidden layer activations, preserving the pairwise distances between points as much as possible in the lower-dimensional space. In particular, MDS is an efficient embedding technique to visualize high-dimensional point clouds by projecting them onto a 2-dimensional plane. Furthermore, MDS has the decisive advantage that it is parameter-free and all mutual distances of the points are preserved, thereby conserving both the global and local structure of the underlying data \cite{torgerson1952multidimensional, kruskal1964nonmetric,kruskal1978multidimensional,cox2008multidimensional, metzner2021sleep, metzner2023extracting, metzner2022classification}. 

When interpreting patterns as points in high-dimensional space and dissimilarities between patterns as distances between corresponding points, MDS is an elegant method to visualize high-dimensional data. By color-coding each projected data point of a data set according to its label, the representation of the data can be visualized as a set of point clusters. For instance, MDS has already been applied to visualize for instance word class distributions of different linguistic corpora \cite{schilling2021analysis}, hidden layer representations (embeddings) of artificial neural networks \cite{schilling2021quantifying, krauss2021analysis}, structure and dynamics of highly recurrent neural networks \cite{krauss2019analysis, krauss2019recurrence, krauss2019weight, metzner2023quantifying}, or brain activity patterns assessed during e.g. pure tone or speech perception \cite{krauss2018statistical, schilling2021analysis}, or even during sleep \cite{krauss2018analysis, traxdorf2019microstructure, metzner2022classification, metzner2023extracting}. 
In all these cases the apparent compactness and mutual overlap of the point clusters permits a qualitative assessment of how well the different classes separate.

By comparing PCA and MDS results, we can assess the robustness of our observed patterns and determine whether apparent clusters or separations are projection-dependent artifacts or genuinely reflect underlying semantic distinctions.

\subsection{Degree of Clustering}

To quantify the degree of clustering, we used the GDV as published and explained in detail in \cite{schilling2021quantifying}. The GDV provides an objective measure of how well the hidden layer activations cluster according to the ASC types, offering insights into the model's internal representations. Briefly, we consider $N$ points $\mathbf{x_{n=1..N}}=(x_{n,1},\cdots,x_{n,D})$, distributed within $D$-dimensional space. A label $l_n$ assigns each point to one of $L$ distinct classes $C_{l=1..L}$. In order to become invariant against scaling and translation, each dimension is separately z-scored and, for later convenience, multiplied with $\frac{1}{2}$:
\begin{align}
s_{n,d}=\frac{1}{2}\cdot\frac{x_{n,d}-\mu_d}{\sigma_d}.
\end{align}
Here, $\mu_d=\frac{1}{N}\sum_{n=1}^{N}x_{n,d}\;$ denotes the mean,\\ \\
and $\sigma_d=\sqrt{\frac{1}{N}\sum_{n=1}^{N}(x_{n,d}-\mu_d)^2}$ the standard deviation of dimension $d$. \\ \\
Based on the re-scaled data points $\mathbf{s_n}=(s_{n,1},\cdots,s_{n,D})$, we calculate the {\em mean intra-class distances} for each class $C_l$ 
\begin{align}
\bar{d}(C_l)=\frac{2}{N_l (N_l\!-\!1)}\sum_{i=1}^{N_l-1}\sum_{j=i+1}^{N_l}{d(\textbf{s}_{i}^{(l)},\textbf{s}_{j}^{(l)})},
\end{align}
and the {\em mean inter-class distances} for each pair of classes $C_l$ and $C_m$
\begin{align}
\bar{d}(C_l,C_m)=\frac{1}{N_l  N_m}\sum_{i=1}^{N_l}\sum_{j=1}^{N_m}{d(\textbf{s}_{i}^{(l)},\textbf{s}_{j}^{(m)})}.
\end{align}
Here, $N_k$ is the number of points in class $k$, and $\textbf{s}_{i}^{(k)}$ is the $i^{th}$ point of class $k$.
The quantity $d(\textbf{a},\textbf{b})$ is the euclidean distance between $\textbf{a}$ and $\textbf{b}$. Finally, the Generalized Discrimination Value (GDV) is calculated from the mean intra-class and inter-class distances  as follows:
\begin{align}
\mbox{GDV}=\frac{1}{\sqrt{D}}\left[\frac{1}{L}\sum_{l=1}^L{\bar{d}(C_l)}\;-\;\frac{2}{L(L\!-\!1)}\sum_{l=1}^{L-1}\sum_{m=l+1}^{L}\bar{d}(C_l,C_m)\right]
 \label{GDVEq}
\end{align}

\noindent whereas the factor $\frac{1}{\sqrt{D}}$ is introduced for dimensionality invariance of the GDV with $D$ as the number of dimensions.

\vspace{0.2cm}\noindent Note that the GDV is invariant with respect to a global scaling or shifting of the data (due to the z-scoring), and also invariant with respect to a permutation of the components in the $N$-dimensional data vectors (because the euclidean distance measure has this symmetry). The GDV is zero for completely overlapping, non-separated clusters, and it becomes more negative as the separation increases. A GDV of -1 signifies already a very strong separation and perfect clustering.

\subsection{Summary}

Our methodology involves creating a dataset of 1000 narratives that combine varying authorial styles and content, encoding these texts through BERT’s layered representations, and projecting them into two-dimensional spaces for interpretability. By applying PCA, MDS, and GDV to these embeddings, we gain insight into how and where within BERT’s representation space different semantics manifest, offering a window into localized processing in deep language models.

\section{Results}
\begin{figure}
    \centering
    \includegraphics[width=\linewidth]{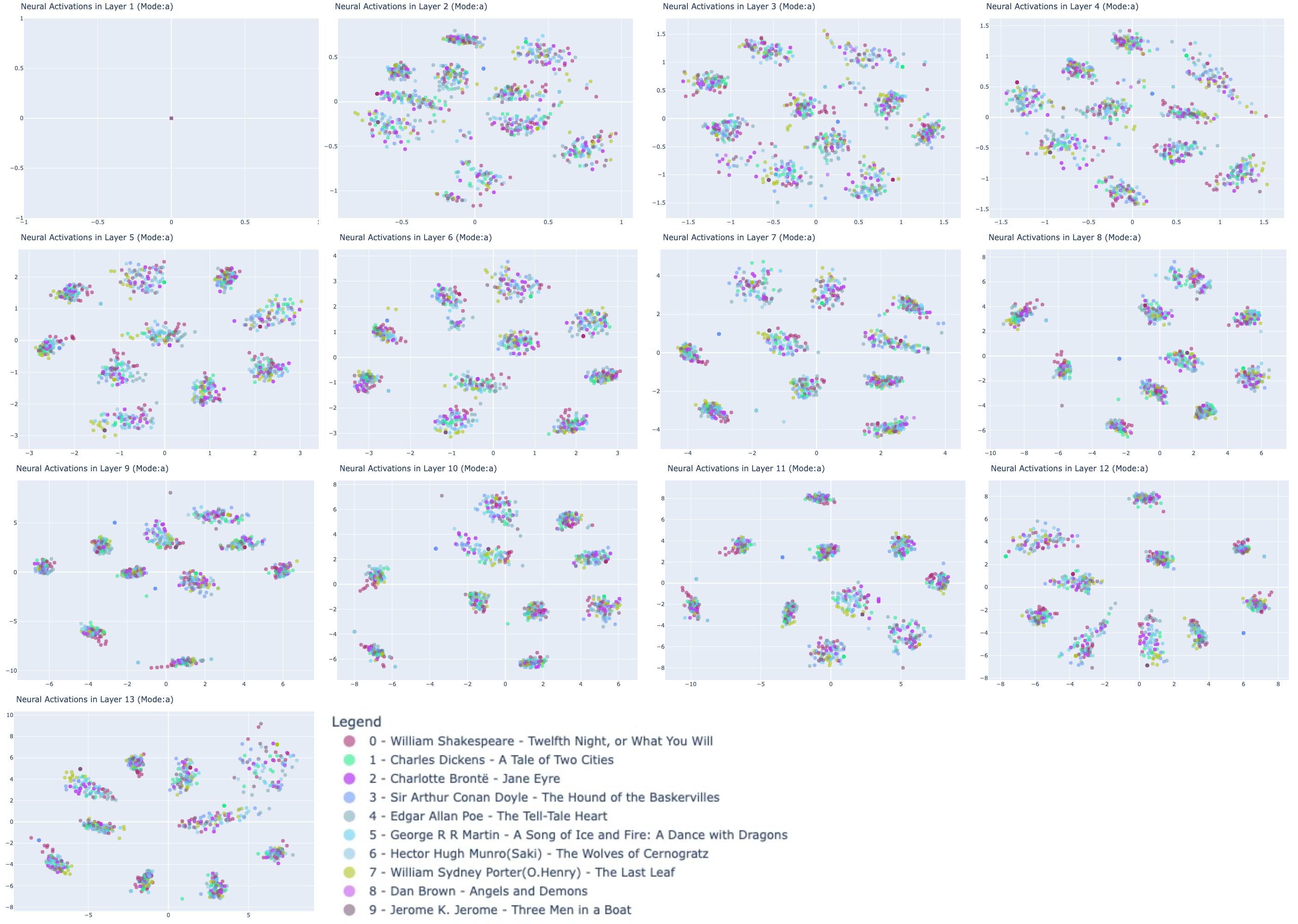} 
    \caption{Layerwise Neural Activations in BERT Projected using MDS when Clustered Based on Authors}
    \label{fig:mds_a}
\end{figure}

\begin{figure}
    \centering
    \includegraphics[width=\linewidth]{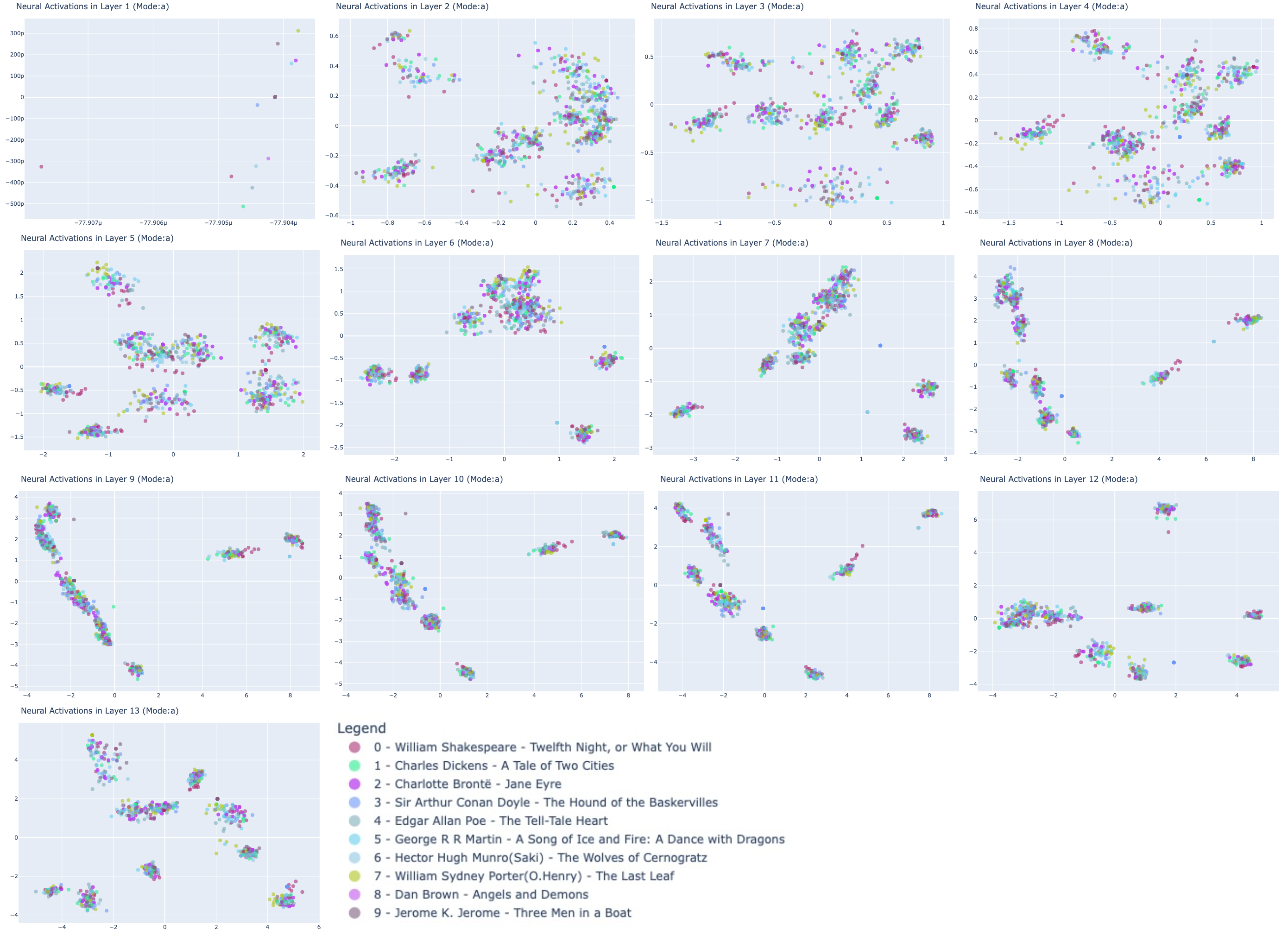} 
    \caption{Layerwise Neural Activations in BERT Projected using PCA when Clustered Based on Authors}
    \label{fig:pca_a}
\end{figure}

\begin{figure}[H]
    \centering
    \includegraphics[width=\linewidth]{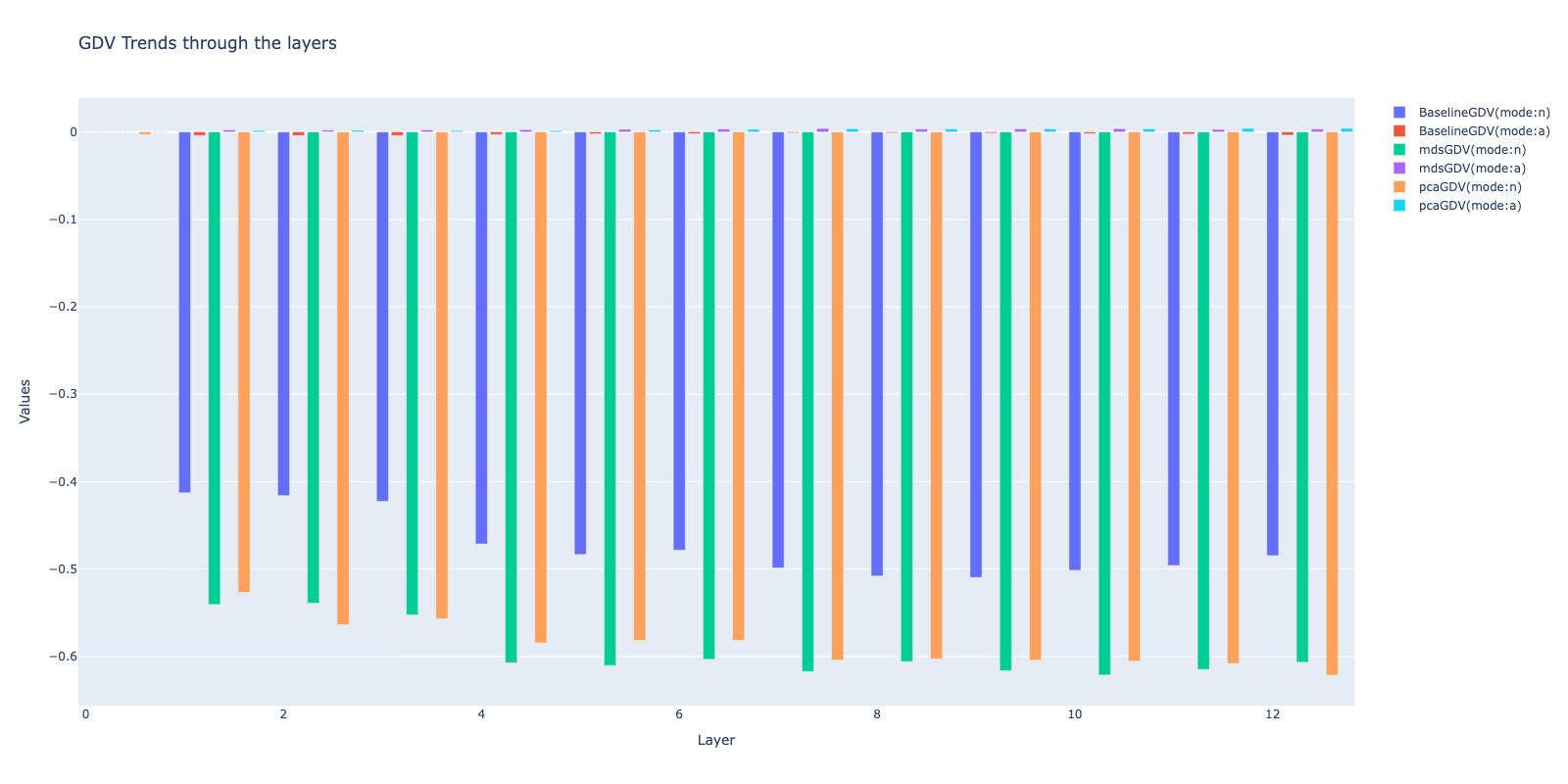} 
    \caption{Layerwise Trends in GDV Illustrating Degree of Clustering (lower is better)}
    \label{fig:gdv}
\end{figure}

From Fig.\ref{fig:mds_a} and Fig.\ref{fig:pca_a}, it is clear that no significant clusters emerge when the data is clustered based on authorial style. This is further supported by Fig.\ref{fig:gdv}, where layerwise GDV trends for authorial style remain close to zero or slightly positive, indicating minimal clustering.

\begin{figure}
    \centering
    \includegraphics[width=\linewidth]{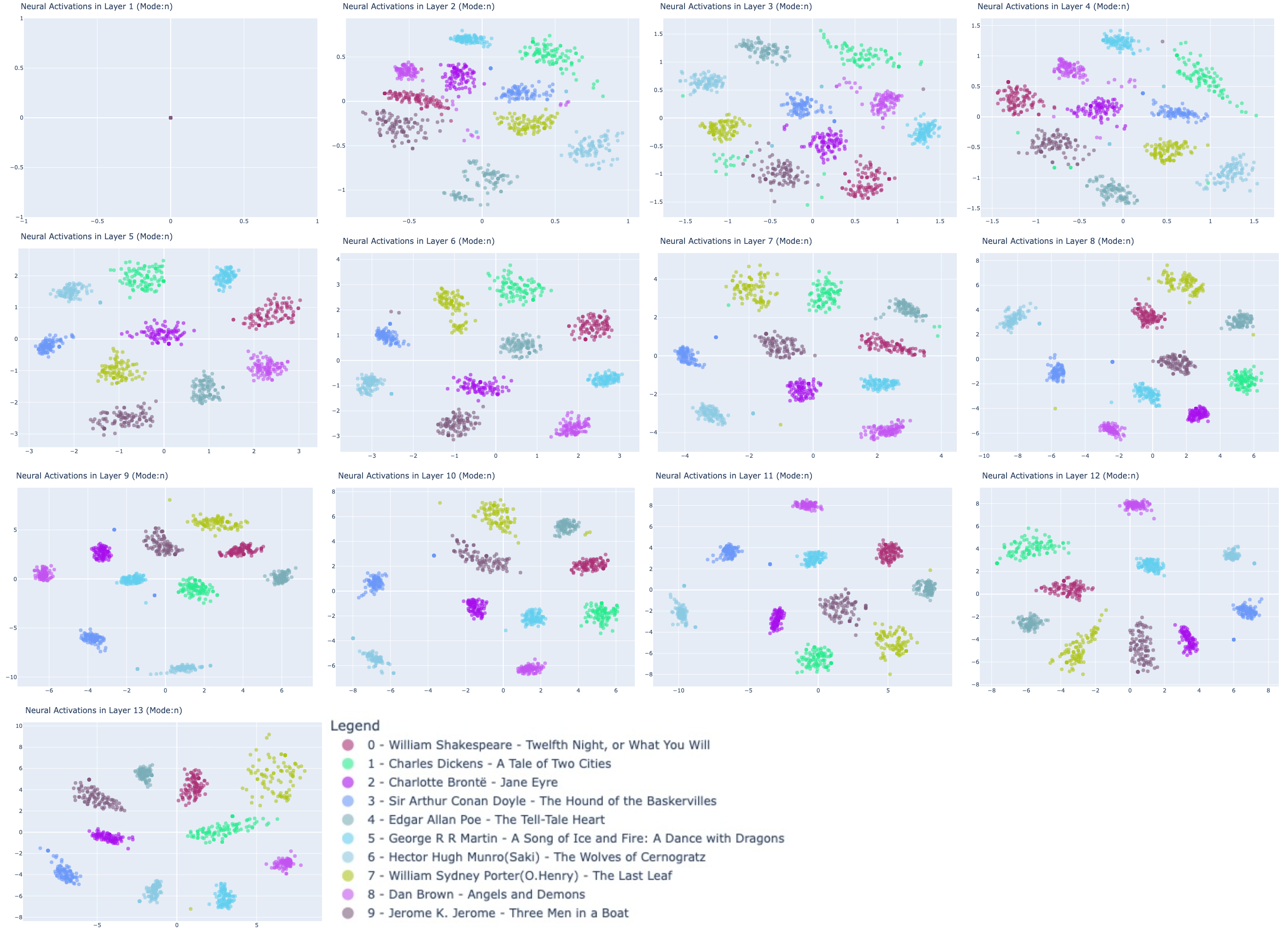} 
    \caption{Layerwise Neural Activations in BERT Projected using MDS when Clustered Based on the Narratives}
    \label{fig:mds_n}
\end{figure}

\begin{figure}
    \centering
    \includegraphics[width=\linewidth]{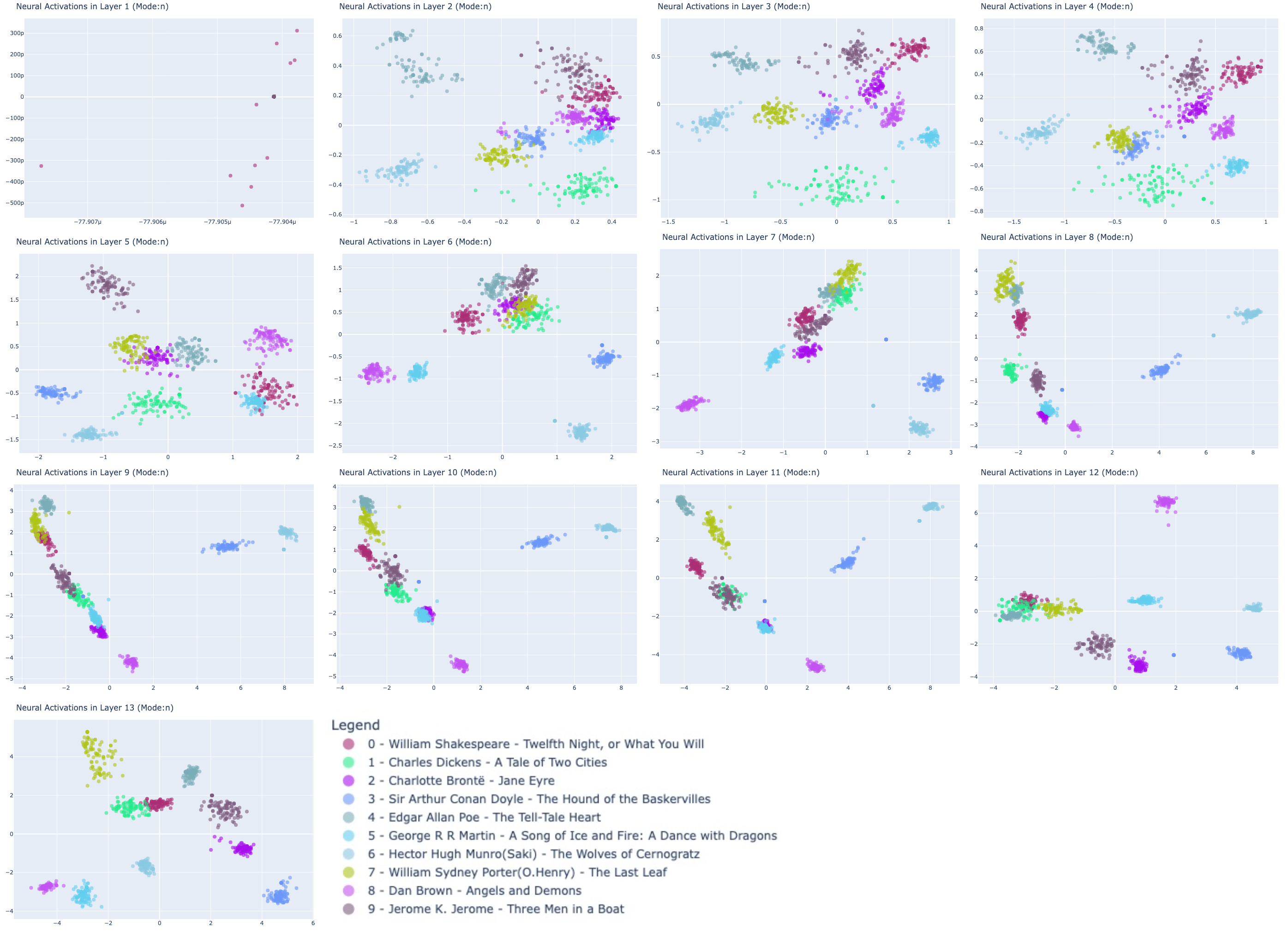} 
    \caption{Layerwise Neural Activations in BERT Projected using PCA when Clustered Based on the Narratives}
    \label{fig:pca_n}
\end{figure}

In contrast, when clustering is based on narrative content, Fig.\ref{fig:mds_n} and Fig.\ref{fig:pca_n} demonstrate clear and compact clusters, with almost no outliers. As shown in Fig.\ref{fig:gdv}, the layerwise GDV trends for narrative content are highly negative, becoming progressively more compact and distinct in the later layers of BERT.

Baseline GDVs calculated on the dataset labeled by narrative content are already negative across layers, as shown in Fig.\ref{fig:gdv}, indicating strong clustering in the 768-dimensional activation space. In contrast, baseline GDVs for authorial style remain close to zero.

These consistent observations validate the projection methodology and confirm that for this dataset, BERT prioritizes narrative content as the primary distinguishing feature, with minimal encoding of authorial style.

\section{Conclusion}
This study explored the hypothesis that localized neural processing, a hallmark of sensory information processing in the human brain, could extend to sequential data and be modeled using artificial neural networks. By analyzing BERT’s layerwise activations in response to a dataset containing texts with distinct authorial styles and narrative content, we uncovered compelling evidence that supports this hypothesis within the capabilities of transformers.

Our results reveal that BERT’s activations exhibit clear clustering based on narrative content across all layers, with progressively compact and distinct clusters in later layers. This finding indicates that BERT prioritizes content as the principal distinguishing feature in its representations, aligning with the hierarchical abstraction of semantic information. Conversely, no meaningful clustering was observed for authorial styles, as reflected in both the projected and raw activation spaces, demonstrating that stylistic attributes are not prominently encoded by BERT.

Interestingly, the observed lack of clustering for authorial styles contrasts with the potential for strong clustering when narratives are rephrased as different text types, such as fables, sci-fi, or children’s stories, as noted in prior studies \cite{krauss2024analyzing}. This distinction suggests that BERT is sensitive to clear stylistic transformations across text types but less so to subtle variations inherent in individual authorial styles. These findings, therefore, highlight the model’s greater emphasis on semantic and structural shifts rather than nuanced stylistic differences.

These findings confirm the ability of encoder-only transformers like BERT to exhibit localized processing of sequential data, drawing an analogy to the brain’s localization of sensory information processing. Furthermore, the consistent layerwise trends validate the projection methodology employed. While not a direct model of biological neural networks, transformers offer valuable abstractions for studying cognitive and sequential processing, paving the way for future interdisciplinary research.

\section{Data and Code Availability Statement}
The complete data and analysis programs will be made
available upon reasonable request.

    
\section*{Acknowledgements}
This work was funded by the Deutsche Forschungsgemeinschaft (DFG, German Research Foundation): KR\,5148/3-1 (project number 510395418), KR\,5148/5-1 (project number 542747151), and GRK\,2839 (project number 468527017) to PK, and grant SCHI\,1482/3-1 (project number 451810794) to AS.



\vspace{12pt}

\end{document}